\pdfoutput=1
\documentclass[runningheads]{llncs}
\usepackage{graphicx}
\usepackage{times}
\usepackage{latexsym}
\usepackage{soul}
\usepackage{amsmath}
\usepackage{multirow}
\usepackage{subcaption}
\usepackage{hyperref}
\usepackage[T1]{fontenc}
\usepackage{xcolor}
\usepackage[utf8]{inputenc}

\usepackage{microtype}

\usepackage{inconsolata}

\begin{document}
\title{Context-Enhanced Language Models for Generating Multi-Paper Citations}
\titlerunning{Context-Enhanced Language Models for Generating Multi-Paper Citations}

% \author{Avinash Anand\orcidID{0009-0003-2479-0342} \and
% Kritarth Prasad\orcidID{0009-0006-2279-7112} \and
% Ujjwal Goel\orcidID{0009-0004-2025-9359} \and
% Mohit Gupta\orcidID{0009-0001-8528-5169} \and
% Naman Lal\orcidID{0009-0008-2914-5509} \and
% Astha Verma\orcidID{0000-0003-3615-5373} \and
% Rajiv Ratn Shah\orcidID{0000-0003-1028-9373}
% }
\author{Avinash Anand \and
Kritarth Prasad \and
Ujjwal Goel \and
Mohit Gupta \and
Naman Lal \and
Astha Verma \and
Rajiv Ratn Shah
}
\institute{Indraprastha Institute of Information Technology, Delhi \\
\email{\{avinasha, kritarth20384, ujjwal20545, mohit22112, asthav, rajivratn\}@iiitd.ac.in}\\
\email{namanlal.lal92@gmail.com}
}
\authorrunning{A. Anand et al.}

\maketitle
\begin{abstract}
Citation text plays a pivotal role in elucidating the connection between scientific documents, demanding an in-depth comprehension of the cited paper. Constructing citations is often time-consuming, requiring researchers to delve into extensive literature and grapple with articulating relevant content. To address this challenge, the field of citation text generation (CTG) has emerged. However, while earlier methods have primarily centered on creating single-sentence citations, practical scenarios frequently necessitate citing multiple papers within a single paragraph. To bridge this gap, we propose a method that leverages Large Language Models (LLMs) to generate multi-citation sentences. Our approach involves a single source paper and a collection of target papers, culminating in a coherent paragraph containing multi-sentence citation text. Furthermore, we introduce a curated dataset named MCG-S2ORC, composed of English-language academic research papers in Computer Science, showcasing multiple citation instances. In our experiments, we evaluate three LLMs LLaMA, Alpaca, and Vicuna to ascertain the most effective model for this endeavor. Additionally, we exhibit enhanced performance by integrating knowledge graphs from target papers into the prompts for generating citation text. This research underscores the potential of harnessing LLMs for citation generation, opening a compelling avenue for exploring the intricate connections between scientific documents.
\end{abstract}

\section{Introduction}
The generation of text in the context of science is a difficult undertaking that calls for a thorough comprehension of the input text and expertise in the relevant field. Citation Generation has received a lot of attention recently because of developments in writing assistants and language models like Transformers \cite{vaswani2017attention}. The Citation Text Generation (CTG) challenge involves generating text that appropriately cites or refers to a cited document in a source document using natural language. The source and the cited paper's contextual signals are frequently used in CTG to generate the text. In this procedure, an algorithmic model should sum up how the original and the cited article relate to one another in a certain situation. This may involve examining the papers content to determine their relationships and applying the appropriate terminology and structure to convey this information clearly and concisely. The time and effort needed for a literature review can be greatly decreased by having the ability to automatically describe the relationships between scientific papers and generate text for these descriptions.

\begin{figure}[ht]
\centering
  \includegraphics[width = 0.9\linewidth]{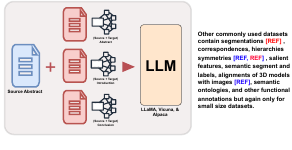}
  \caption{Multi-Sentence Citation Text Generation}
\label{fig:multi_text_citation_dataset}
\end{figure}

However, there are a few drawbacks to the present citation text generation technologies. They concentrated on coming up with a single sentence for a single reference~\cite{cohan2019structural}. 

The text from the abstract of papers was mostly used as input in earlier attempts to CTG~\cite{Luu2020CitationTG} methods, which produced a single citation as an output. In real-world scenerios, authors often use multiple references in one sentence or paragraph. Wu et al.~\cite{wu2021towards} initiative suggests a method for producing multi-reference citation text using the source and target abstracts as input. They obtained the sentences for their dataset of multiple reference citations from the ACL anthology. For the task of creating citation text, they employed the Fusion-in-Decoder (FiD)~\cite{izacard-grave-2021-leveraging} model. They have used FiD because it can scan lengthy inputs from various publications and makes use of the generational power of huge pre-trained contextual language models like T5~\cite{JMLR:v21:20-074}. They have incorporated the intent labels to improve the performance of the model. 

Using a CTG model offers numerous advantages. Firstly, it significantly saves time by automating the citation generation process, allowing researchers, students, and authors to focus more on their work. Secondly, these models ensure accuracy and consistency by following specific citation styles and providing correct formatting for various source types. Thirdly, citation generation models contribute to reducing plagiarism by encouraging proper source attribution. Moreover, these models offer flexibility by supporting multiple citation styles to accommodate different disciplines. Lastly, CTG models can serve as educational resources, helping students learn about citation elements and proper practices. Overall, CTG models streamline the citation process, improve accuracy and adherence to styles, save time, and promote ethical writing practices.

Our research presents a way to produce multi-sentence citation text for the target and source abstracts that are provided by using large language models. We have fine-tuned three LLMs i.e. LLaMA~\cite{LLaMA}, Alpaca~\cite{alpaca}, and Vicuna~\cite{vicuna2023}. The Fig.~\ref{fig:multi_text_citation_dataset} shows the basic workflow of our approach, and defines our citation generation pipeline. We have proposed a new dataset MCG-S2ORC, which is created from the S2ORC~\cite{lo2019s2orc} dataset for the task of multi-citation. Additionally, we demonstrate that by including knowledge graphs of the source and target papers in the prompts improves the performance of our model. The knowledge graphs relations are extracted from the abstracts of research papers~\cite{kgfromabstracts}, which contain condensed information and dependencies between the phrases. The relations are extracted with the help of \textbf{PL-Marker}~\cite{ye2022plmarker}. We have shown that the LLMs performs better for the multi-reference CTG challenge by integrating these prompts. We offer the following summary of our contributions:

\begin{itemize}
\item We propose a Citation Generation architecture that takes the
abstract of papers as input for the CTG task. Additionally, we incorporated the knowledge graphs in the prompts for the LLMs to show improved performance over the baselines.
\item We propose MCG-S2ORC dataset using the S2ORC~\cite{lo2019s2orc} dataset. This dataset contains two-three target papers for a single source paper. 
\item We show the importance of incorporating knowledge graph in the prompt structure for LLM through a huge increase in performance.
\end{itemize}

The written work is structured as follows: Section \ref{literature} addresses the related works on citation text generation, Section \ref{methodology} explains how we came up with the problem and how the dataset was created, \& what models have been used. In Section~\ref{experiments}, we have explained about how we performed our experiments, then the Section~\ref{discussion} shows the evaluations, and the paper's conclusion and future aims are summarised in Section \ref{conclusion}.

\section{Related Work\label{literature}}
\subsection{Text Generation}
Koncel et al.~\cite{koncel2019text} generated multi-sentence text from an information extraction system and improved performance using a knowledge graph. They did graph encoding using Graph Attention Network. Text generation for scientific documents is one example of multi-document scientific summarization, other examples include the task of multi-document scientific summarization in the scientific domain ~\cite{chen2014summarization,mohammad2009using,yeloglu2011multi}. Chen et al.~\cite{chen2021scixgen} proposed a SciXGen dataset to solve the problem of generation of context-aware text in a scientific domain. However, summarising academic papers differs from the CTG job.

\subsection{Citation Text Generation}
As far as we are aware, there are two active concurrent works~\cite{Luu2020CitationTG,xing2020automatic} that generate citation texts from research papers. Luu et al.~\cite{Luu2020CitationTG} first introduced the task and generated the citation text given source and cited documents. Xing et al.~\cite{xing2020automatic} explored the relationship between scientific documents on a larger dataset.  Gu and Hahnloser~\cite{gu2022controllable} proposes a pipeline for controllable citation generation that consists of an attribute recommendation module and a module for conditional citation generation, and evaluates the system's controllability across numerous characteristics using both automated metrics and human review. Jung et al.~\cite{intentControllable} proposes a framework for controllable citation generation with three labels of intent background, method, and results. They have used BART and T5 transformer and compares the results and accuracies obtained using these 2 transformer-based models. 

There has been very less work done on multi-reference citation text generation, and we have found~\cite{wu2021towards} in which authors concentrate on generating multiple citations from the sources and cited papers. To handle diverse long inputs, they create a new generation model using the Fusion-in-Decoder method.

\subsection{Large Language Models \& Prompts}
The emergence of large language models (LLMs) has been a significant advancement in the field of education~\cite{10.1007/978-3-031-49601-1_5,10.1007/978-3-031-49601-1_4,mathify}, as well as in CTG tasks, as demonstrated by recent studies such as those conducted by Avinash et al.~\cite{10.1007/978-3-031-49601-1_3}. These versatile models have opened up a plethora of new learning opportunities. Ye et al.~\cite{ye2023incontext}, In-context instruction learning, combines instruction prompting and few-shot learning. The prompt includes a number of demonstration examples for various tasks, with each demonstration including an instruction, task input, and task output. Used in Stanford Alpaca~\cite{alpaca} to generate 52k instructions following text-davinci-003 GPT3~\cite{gpt3paper} prompts and then fine-tune LLaMA. Chain-of-thought (CoT) prompting~\cite{wei2023chainofthought} generates a sequence of short sentences to describe reasoning logic step by step.

\section{Methodology\label{methodology}}
In this paper, we fine-tuned three large language models (LLMs), i.e. LLaMA~\cite{LLaMA}, Alpaca~\cite{alpaca}, and Vicuna~\cite{vicuna2023} for the task of generating Multi-citation text. All the models were evaluated using the metrics METEOR, Rouge-1, Rouge-2, and Rouge-L. These fine-tuned models are considered as our baselines. We then extracted the relations from the source and target papers and use them in the prompt for generating the citations. Based on our empirical findings, we observed that incorporating knowledge graph relations in the prompting process enhances the performance of generating citation texts when compared to our baseline models.

\subsection{Problem Formulation \& Notations}
The problem statement includes: given a abstract of citing document $A$, set of abstracts, introductions, and conclusions of related documents $B = \{b_1, b_2, ..., b_n\}$. The task aims to produce a multi-sentence paragraph of all the cited documents $b_i$ in the context of citing abstract $A$. We curated our own dataset from the benchmark dataset S2ORC~\cite{lo2019s2orc}. We have modified the dataset in such a way that for each citing abstract, we have added more two-three cited papers, and the target has multiple sentences with multiple references for each pair $(A, B)$. Fig.~\ref{fig:multi_text_citation_dataset} clearly demonstrate how our approach is going to work. First we take the source abstract, then we add them with the abstract, introduction, and conclusion of the target paper, and extract knowledge graph relation using PL-Marker~\cite{ye2022plmarker}, then pass it in the prompt for LLM, whose structure is given in the Fig.~\ref{kg_prompt}.

\subsection{Dataset}
For the task of multi-sentence citation text generation, we synthesize a new dataset MCG-S2ORC from S2ORC~\cite{lo2019s2orc}. The S2ORC\footnote{\url{https://github.com/allenai/s2orc}}, or Semantic Scholar Open Research Corpus, is a significant corpus of 81.1 million English-language academic papers from many academic disciplines. We have taken only the samples whose “Field of Study" contain “Computer Science". Presently, the computer science domain contains 6.0M total papers. Each sample from the set of 6.0M computer science domain papers contains $``source\_paper\_id"$, $``source\_abstract"$, and $``body\_text"$. The $``body\_text"$ consists of various sections, including Introduction, Methodology, etc.

We parsed the Computer Science domain dataset and extracted citation details in JSON format. The dataset consists of samples representing citation examples, each containing key-value pairs of information. The $``source\_paper\_id"$ field provides a unique identifier for the source paper, while the $``source\_abstract"$ field contains its summary. The $``citation\_texts"$ field is an array containing citation information related to the source paper. Each citation includes the $``citation\_text"$ field, representing the extracted citation text. Additional metadata is found in the $``citation\_meta"$ field, containing information like citation number, referenced section, and details about the paper being cited (title, abstract, introduction, and conclusion). To ensure suitability for multi-reference citation text generation, we only considered citations that cite more than one paper in a single sentence, making necessary modifications to the dataset.

Our final dataset comprises 17,210 samples of multi-reference citation texts. The complete statistics of our MCG-S2ORC are shown in Table~\ref{tab:dataset}.

\begin{table*}[ht]
    \centering
    \setlength{\tabcolsep}{5\tabcolsep}
    \begin{tabular}{l|c|c|c|c}
     \hline
     \hline
     \textbf{Statistic} & \textbf{CTG-S2ORC} & \textbf{Train} & \textbf{Validation} & \textbf{Test}\\
     \hline
     \# citations & 17210 & 13,779 & 1,716 & 1,715\\
     \hline
     \# unique papers & 17210 & 13,779 & 1,716 & 1,715\\
     \hline
     \multicolumn{5}{c}{\textbf{CITATIONS}} \\
     \hline
     Avg \# characters & 227.29 & 227.40 & 230.25 & 223.37\\
     \hline
     Max \# characters & 2416 & 2416 & 1862 & 1061\\
     \hline
     % Min \# characters & 1 & 3 & 1 & 2\\
     % \hline
     \multicolumn{5}{c}{\textbf{SOURCE ABSTRACTS}} \\
     \hline
     Avg \# characters & 1122.95 & 1,120.73 & 1,111.55 & 1152.23\\
     \hline
     Max \# characters & 5516 & 5516 & 4343 & 3642\\
     \hline
     % Min \# characters & 1 & 1 & 3 & 3\\
     % \hline
     \multicolumn{5}{c}{\textbf{TARGET ABSTRACTS}} \\
     \hline
     Avg \# characters & 998.48 & 997.87 & 999.35 & 1002.56\\
     \hline
     Max \# characters & 93551 & 93551 & 8674 & 4924\\
     \hline
     % Min \# characters & 2 & 2 & 2 & 3\\
     % \hline
     Avg \# of Targets per sample & 2 & 2 & 2 & 2\\
     \hline
     \hline
    \end{tabular}
    \vspace{0.2cm}
    \caption{Dataset statistics created from the S2ORC corpus.}
\label{tab:dataset}
\end{table*}

\subsection{Large Language Models}
\begin{figure}[ht]
\centering
  \includegraphics[width = 0.5\linewidth]{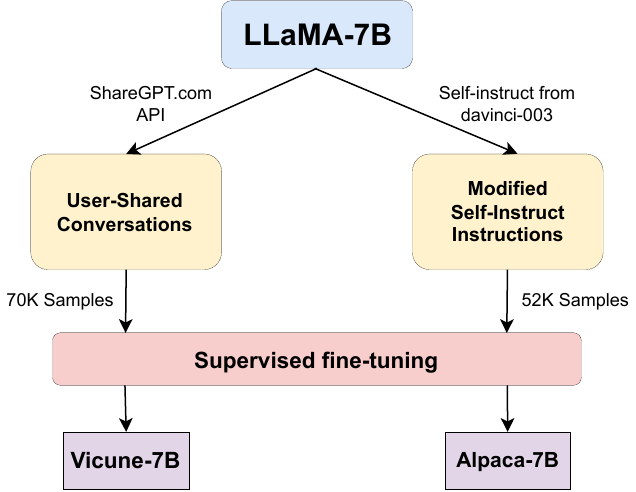}
  \caption{LLaMA, Vicuna \& Alpaca}
\label{fig:large_lang_models}
\end{figure}

We have fine-tuned three large language models for the task of generating multi-sentence citation text. The details of three models can be seen in Fig.~\ref{fig:large_lang_models} and provided below.

\textbf{LLaMA} is a transformer-based model available in four variations: 7B, 13B, 33B, and 65B parameters. Trained solely on publicly available data, the training corpus comprises approximately 1.4T tokens and includes text from 20 different languages~\cite{LLaMA}. \textbf{Alpaca}~\cite{alpaca}, on the other hand, is a language model that has undergone supervised fine-tuning using an LLaMA 7B model and 52K instruction-following demonstrations generated by OpenAI's text-davinci-003 model~\cite{LLaMA,gpt3paper}. \textbf{Vicuna}, another variant, has been developed by optimizing an LLaMA base model with approximately 70K user-shared talks obtained from ShareGPT.com via open APIs~\cite{vicuna2023}. It is important to note that Vicuna is limited in its reasoning abilities, mathematical understanding, self-identification capabilities, fact-checking capacity, and it is not specifically optimized for bias reduction, potential toxicity, or safety measures~\cite{vicuna2023}.

\begin{figure}
\centering
\begin{subfigure}{.5\textwidth}
\centering  \includegraphics[width=.7\linewidth]{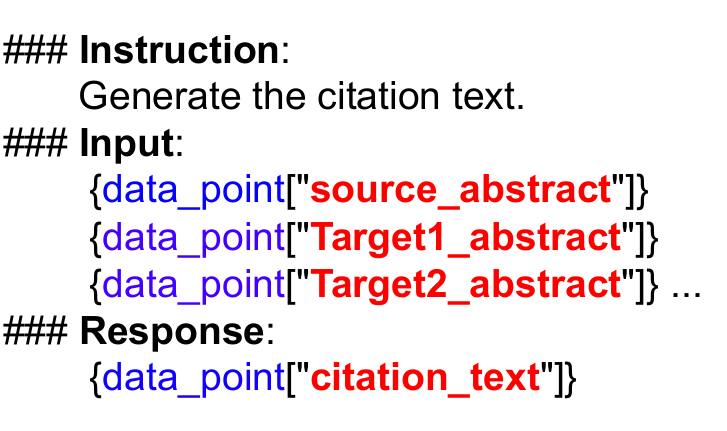}
  \caption{}
  \label{prompt_structure}
\end{subfigure}%
\begin{subfigure}{.5\textwidth}
  \centering
\includegraphics[width=.9\linewidth]{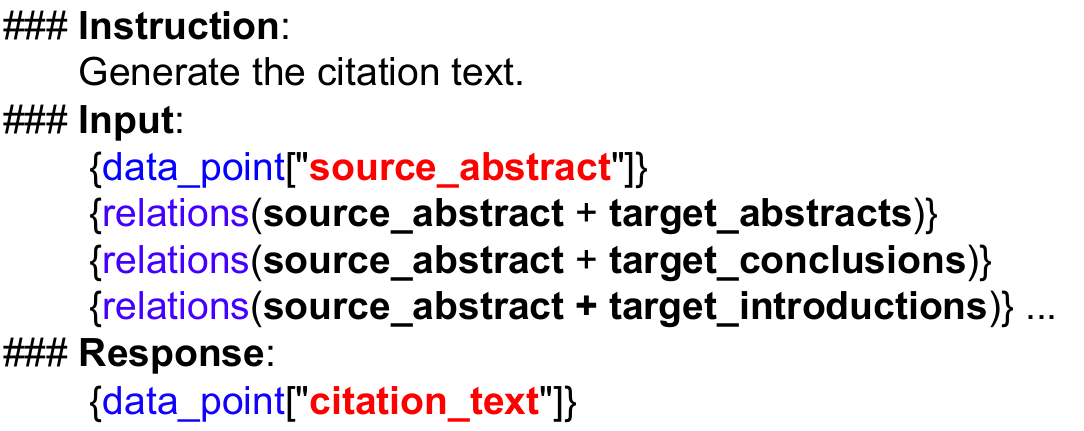}
  \caption{}
  \label{kg_prompt}
\end{subfigure}
\caption{Prompt Structures used for the Large Language Models.}
\label{Prompts}
\end{figure}

The prompt used to fine-tune the baselines LLMs are provided in the Fig.~\ref{prompt_structure}. In the figure, the \textbf{data\_point} is a dictionary containing a single data sample from our dataset.

% \begin{figure}[h]
% \centering
%   \includegraphics[width = 6cm]{prompt_structure_without.pdf}
%   \caption{The Prompt Structure}
% \label{prompt_structure}
% \end{figure}

\subsection{Prompting \& Knowledge Graphs}
We also attempted adding knowledge graph relations of the abstract, introduction, and conclusion of the target paper and abstract of source paper in the prompts for fine-tuning the LLM models, LLaMA, Alpaca, and Vicuna, which shows a huge improvement in the results, as using relations of paper in the prompts in generating outputs from large language models provides a specific instruction or context to guide the model's response allowing more focused and relevant output, it also enables control over the style, tone, or domain of the generated text. 

Adding knowledge graph relations to prompts for text generation has several advantages. Firstly, it improves contextual understanding by enabling the model to comprehend entity relationships, enhancing its grasp of the topic. Secondly, it contributes to enhanced coherence and consistency in the generated text. By leveraging graph relationships, the model produces more structured and coherent responses, improving overall quality. Additionally, the integration of knowledge graphs allows the model to showcase domain-specific expertise, delivering informed and accurate responses. Lastly, knowledge graphs aid in fact-checking and verification, ensuring factual accuracy and reducing the likelihood of generating misleading information.

In this work, we utilized the PL-Marker~\cite{ye2022plmarker} tool to construct the knowledge graph of the source and target abstracts. PL-Marker employs an innovative packed levitated marker technique, combining both a neighborhood-oriented and subject-oriented packing strategy to obtain pair representations. The purpose of constructing the knowledge graph is to capture the relationships and context between different entities within the abstracts of papers. The first step of the model involves entity recognition, where it identifies and labels the different entities present in the text. Once the entities are recognized and labelled, the model focuses on extracting relations between these entities. We generated knowledge graph triplets for target paper's introduction, conclusion and abstract with the abstract of source paper which are used in the dataset to fine-tune the LLM’s. The prompt structure is given in the Fig.~\ref{kg_prompt}.

% \begin{figure}[h]
% \centering
%   \includegraphics[width = 6cm]{prompt_structure_with_kg.pdf}
%   \caption{The Prompt Structure with Relations}
% \label{kg_prompt}
% \end{figure}

\begin{figure}[ht]
\centering
  \includegraphics[width = 0.9\linewidth]{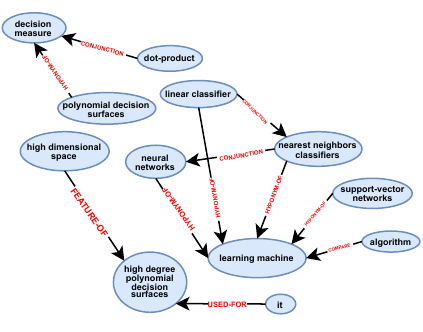}
  \caption{Knowledge Graph Visualization}
\label{knowledge_graph_relations}
\end{figure}

The example visualization of extracted relations from the target introduction that we have used in the prompts is shown in Fig.~\ref{knowledge_graph_relations}. This figure clearly shows that the our approach is able to extract complex relationship between different tokens.

\section{Experiments\label{experiments}}
We split the complete dataset MCG-S2ORC containing 17,210 data samples into train, test, and validation set having 13K, 1K, and 1K samples respectively. After creating and preprocessing the dataset, we fine-tuned three large language models as discussed earlier on our dataset MCG\_S2ORC for citation text generation. The prompt for the LLMs as shown in Fig.~\ref{prompt_structure} is converted into tokens, then pass it to the model for fine-tuning and learning the weights. From the results shown in Table~\ref{tab:result_table}, \textbf{Vicuna}~\cite{vicuna2023} outperforms LLaMA and alpaca for the task of citation text generation on our dataset. These results act as our baselines for our next setup.

Then we further perform experiments on passing knowledge graph of the abstract, introduction, and conclusion of target paper as prompts to better capture the relationship and coherence of the words to generate more meaningful citations. We have extracted the knowledge graph relations using PL-Marker~\cite{ye2022plmarker}. The results for this setup can be seen from Table~\ref{tab:table_result_with_prompt}, the performance of all the models is improved from our baselines. 

For fine-tuning the Large Language Models (LLMs), we employed QLora~\cite{dettmers2023qlora}. QLora is an efficient approach that maximize memory efficiency through gradient backpropagating gradients in a frozen, 4-bit quantized pretrained language model, resulting in Low Rank Adapters (LoRA).
\begin{equation}
    k_i = \frac{1}{2}\left(Q_X\left(\frac{i}{2^n + 1}\right) + Q_X\left(\frac{i+1}{2^n + 1}\right)\right)
\end{equation}
Where, $Q_x (.)$ is the quantile function of the standard normal distribution $N(0, 1)$. For our experiments, we have used $n = 4$ as we are applying 4-bit quantization.
\begin{equation}
    \begin{split}
        m_t = \beta m_{t-1} + \eta \nabla J(w_t)\\
        v_t = \gamma v_{t-1} + (1 - \gamma) \nabla J(w_t)^2
    \end{split}
\label{equation_momentum}
\end{equation}

We utilized the \textbf{AdamW} optimizer~\cite{kingma2014adam} with a Linear Scheduler. The learning rate was set to 3e-4, and we incorporated 100 warmup steps to gradually adjust the learning rate.
\begin{align}
    \hat{m_t} = \frac{m_t}{1 - \beta^t} && \hat{v_t} = \frac{v_t}{1 - \gamma^t}
\label{bias_correction}
\end{align}

Equation.~\ref{bias_correction} shows the bias correction, then the final weight update equation for the Adam optimizer is given by:
\begin{equation}
    w_{t+1} = w_{t} - \frac{\eta}{\sqrt{\hat{v}_t + \epsilon}}\hat{m}_t
\end{equation}

where $\epsilon$ is the error term, which is used such that denominator never reaches zero.  

\subsection{Evaluation Metrics}
The three models results were compared to assess the effectiveness of the generated citation text. The degree of similarity between the generated and actual reference citation texts in the citing paper served as a measure of performance.

We evaluated the generated citation text using standard text creation and summarization metrics: METEOR, ROUGE-N, and ROUGE-L. METEOR combines precision, recall, and alignment-based measures to assess the similarity between the generated citation text and the original reference citation texts. ROUGE-L specifically focuses on the longest common subsequence (LCS) between the generated and reference texts, evaluating the fluency and coherence of the generated text. ROUGE-N extends this evaluation to consider n-gram overlaps, providing a more detailed analysis of the generated text's performance.

\section{Results \& Discussion\label{discussion}}
The results of the experiments after fine-tuning the LLM models for multi-reference citation text generation can be seen from table~\ref{tab:result_table}. The results shows that the Vicuna~\cite{vicuna2023} outperforms other models with respect to all the metrics. The citation text generated by the best fine-tuned model Vicuna is given in the appendix Figure.~\ref{example1}.

\begin{table*}[ht]
    \centering
    \setlength{\tabcolsep}{12\tabcolsep}
    \begin{tabular}{|c|c|c|c|c|}
     \hline
     \textbf{Model} & \textbf{METEOR} & \textbf{Rouge-1} & \textbf{Rouge-2} & \textbf{Rouge-L} \\
     \hline
     \hline
     \textbf{LLaMA} & 11.73 & 10.74 & 1.21 & 9.15 \\
     \hline
     \textbf{Alpaca} & 9.74 & 9.04 & 1.33 & 7.78 \\
     \hline
     \textbf{Vicuna} & \textbf{12.56} & \textbf{12.02} & \textbf{1.44} & \textbf{10.24} \\
     \hline
     \end{tabular}
     \vspace{0.2cm}
    \caption{Results of Fine-Tuned LLM}
\label{tab:result_table}
\end{table*}

Table~\ref{tab:table_result_with_prompt} presents the evaluation results obtained by incorporating knowledge graph relations of source and target paper's abstract, introduction, and conclusion in the prompts. The findings highlight the superior performance of Vicuna, surpassing other models in the specific task of citation generation as compared to the baseline models without knowledge graph relations. This notable achievement can be attributed to the utilization of knowledge graphs, which facilitate a deeper contextual comprehension and enhance the coherence of the generated text. Consequently, the model produces outputs that are more context-rich and of higher quality, ultimately contributing to improved overall performance. The generated citation at the time of inference is given in the Figure.~\ref{example2} in the appendix section.

\begin{table}[ht]
    \centering
    \setlength{\tabcolsep}{12\tabcolsep}
    \begin{tabular}{|c|c|c|c|c|}
     \hline
     \textbf{Model} & \textbf{METEOR} & \textbf{Rouge-1} & \textbf{Rouge-2} & \textbf{Rouge-L} \\
     \hline
     \hline
     \textbf{LLaMA} & 11.46 & 10.79 & 1.23 & 9.14 \\
     \hline
     \textbf{Alpaca} & \textbf{13.39} & 12.42 & \textbf{1.74} & 10.59 \\
     \hline
     \textbf{Vicuna} & 13.18 & \textbf{12.65} & 1.49 & \textbf{10.80} \\
     \hline
     \end{tabular}
     \vspace{0.2cm}
    \caption{Results of Fine-Tuned Model + Knowledge Graph as Prompt}
\label{tab:table_result_with_prompt}
\end{table}

\section{Conclusion\label{conclusion}}
The paper addresses the problem of multi-citation text generation, focusing on generating coherent multi-sentence citations. We curated a dataset called MCG-S2ORC from the S2ORC dataset to advance citation generation research. Three large language models, namely LLaMA, Alpaca, and Vicuna were fine-tuned specifically for citation generation. Vicuna demonstrated superior performance compared to the other models. To enhance citation generation, we integrated knowledge graphs into the model's prompts by extracting entity relations from the source and target paper's abstracts, introductions, and conclusions using PL-Marker. Our experiments showed that incorporating knowledge graphs significantly improved the performance and text generation capabilities of the models, enabling better comprehension of relations between source and target papers. This integration enhances the citation generation task, showcasing the potential of knowledge graphs as valuable resources. Future research can leverage knowledge graphs to explore novel approaches for generating accurate and coherent multi-sentence citations.

\section{Limitations\label{limitations}}
The maximum token length restriction of the LLMs used, set at 2048, is one restriction on our approach. This restricts us to incorporating only 2-3 combined relations between source and target papers rather than including all target papers. While including all sets of relations could enhance performance, it presents challenges due to the increased number of tokens involved.

\section{Acknowledgements}
Rajiv Ratn Shah is partly supported by the Infosys Center for AI, the Center for Design and New Media, and the Center of Excellence in Healthcare at IIIT Delhi.

\bibliographystyle{splncs04}
\bibliography{custom}

\begin{thebibliography}{10}
\providecommand{\url}[1]{\texttt{#1}}
\providecommand{\urlprefix}{URL }
\providecommand{\doi}[1]{https://doi.org/#1}

\bibitem{10.1007/978-3-031-49601-1_5}
Anand, A., Addala, K., Baghel, K., Goel, A., Hira, M., Gupta, R., Shah, R.R.:
  Revolutionizing high school physics education: A novel dataset. In: Big Data
  and Artificial Intelligence: 11th International Conference, BDA 2023, Delhi,
  India, December 7–9, 2023, Proceedings. p. 64–79. Springer-Verlag,
  Berlin, Heidelberg (2023), \url{https://doi.org/10.1007/978-3-031-49601-1_5}

\bibitem{10.1007/978-3-031-49601-1_4}
Anand, A., Goel, A., Hira, M., Buldeo, S., Kumar, J., Verma, A., Gupta, R.,
  Shah, R.R.: Sciphyrag - retrieval augmentation to improve llms on physics
  q\&a. In: Big Data and Artificial Intelligence: 11th International
  Conference, BDA 2023, Delhi, India, December 7–9, 2023, Proceedings. p.
  50–63. Springer-Verlag, Berlin, Heidelberg (2023),
  \url{https://doi.org/10.1007/978-3-031-49601-1_4}

\bibitem{10.1007/978-3-031-49601-1_3}
Anand, A., Gupta, M., Prasad, K., Goel, U., Lal, N., Verma, A., Shah, R.R.:
  Kg-ctg: Citation generation through knowledge graph-guided large language
  models. In: Big Data and Artificial Intelligence: 11th International
  Conference, BDA 2023, Delhi, India, December 7–9, 2023, Proceedings. p.
  37–49. Springer-Verlag, Berlin, Heidelberg (2023),
  \url{https://doi.org/10.1007/978-3-031-49601-1_3}

\bibitem{mathify}
Anand, A., Gupta, M., Prasad, K., Singla, N., Sanjeev, S., Kumar, J., Shivam,
  A.R., Shah, R.R.: Mathify: Evaluating large language models on mathematical
  problem solving tasks (2023)

\bibitem{chen2021scixgen}
Chen, H., Takamura, H., Nakayama, H.: Scixgen: A scientific paper dataset for
  context-aware text generation. arXiv preprint arXiv:2110.10774  (2021)

\bibitem{chen2014summarization}
Chen, J., Zhuge, H.: Summarization of scientific documents by detecting common
  facts in citations. Future Generation Computer Systems  \textbf{32},
  246--252 (2014)

\bibitem{vicuna2023}
Chiang, W.L., Li, Z., Lin, Z., Sheng, Y., Wu, Z., Zhang, H., Zheng, L., Zhuang,
  S., Zhuang, Y., Gonzalez, J.E., Stoica, I., Xing, E.P.: Vicuna: An
  open-source chatbot impressing gpt-4 with 90\%* chatgpt quality (March 2023),
  \url{https://lmsys.org/blog/2023-03-30-vicuna/}

\bibitem{cohan2019structural}
Cohan, A., Ammar, W., Van~Zuylen, M., Cady, F.: Structural scaffolds for
  citation intent classification in scientific publications. arXiv preprint
  arXiv:1904.01608  (2019)

\bibitem{dettmers2023qlora}
Dettmers, T., Pagnoni, A., Holtzman, A., Zettlemoyer, L.: Qlora: Efficient
  finetuning of quantized llms. arXiv preprint arXiv:2305.14314  (2023)

\bibitem{gu2022controllable}
Gu, N., Hahnloser, R.H.R.: Controllable citation text generation (2022)

\bibitem{izacard-grave-2021-leveraging}
Izacard, G., Grave, E.: Leveraging passage retrieval with generative models for
  open domain question answering. In: Proceedings of the 16th Conference of the
  European Chapter of the Association for Computational Linguistics: Main
  Volume. pp. 874--880. Association for Computational Linguistics, Online (Apr
  2021). \doi{10.18653/v1/2021.eacl-main.74},
  \url{https://aclanthology.org/2021.eacl-main.74}

\bibitem{intentControllable}
Jung, S.Y., Lin, T.H., Liao, C.H., Yuan, S.M., Sun, C.T.: Intent-controllable
  citation text generation. Mathematics  \textbf{10}, ~1763 (05 2022).
  \doi{10.3390/math10101763}

\bibitem{gpt3paper}
Katar, O., Ozkan, D., GPT, Yildirim, O., Acharya, U.R.: Evaluation of gpt-3 ai
  language model in research paper writing (12 2022).
  \doi{10.13140/RG.2.2.11949.15844}

\bibitem{kingma2014adam}
Kingma, D.P., Ba, J.: Adam: A method for stochastic optimization. arXiv
  preprint arXiv:1412.6980  (2014)

\bibitem{koncel2019text}
Koncel-Kedziorski, R., Bekal, D., Luan, Y., Lapata, M., Hajishirzi, H.: Text
  generation from knowledge graphs with graph transformers. arXiv preprint
  arXiv:1904.02342  (2019)

\bibitem{lo2019s2orc}
Lo, K., Wang, L.L., Neumann, M., Kinney, R., Weld, D.S.: S2orc: The semantic
  scholar open research corpus. arXiv preprint arXiv:1911.02782  (2019)

\bibitem{Luu2020CitationTG}
Luu, K., Koncel-Kedziorski, R., Lo, K., Cachola, I., Smith, N.A.: Citation text
  generation. ArXiv  \textbf{abs/2002.00317} (2020)

\bibitem{mohammad2009using}
Mohammad, S., Dorr, B., Egan, M., Hassan, A., Muthukrishnan, P., Qazvinian, V.,
  Radev, D., Zajic, D.: Using citations to generate surveys of scientific
  paradigms. In: Proceedings of human language technologies: The 2009 annual
  conference of the North American chapter of the association for computational
  linguistics. pp. 584--592 (2009)

\bibitem{JMLR:v21:20-074}
Raffel, C., Shazeer, N., Roberts, A., Lee, K., Narang, S., Matena, M., Zhou,
  Y., Li, W., Liu, P.J.: Exploring the limits of transfer learning with a
  unified text-to-text transformer. Journal of Machine Learning Research
  \textbf{21}(140),  1--67 (2020), \url{http://jmlr.org/papers/v21/20-074.html}

\bibitem{kgfromabstracts}
Sun, K., Qiu, Z., Salinas, A., Huang, Y., Lee, D.H., Benjamin, D., Morstatter,
  F., Ren, X., Lerman, K., Pujara, J.: Assessing scientific research papers
  with knowledge graphs. In: Proceedings of the 45th International ACM SIGIR
  Conference on Research and Development in Information Retrieval. p.
  2467–2472. SIGIR '22, Association for Computing Machinery, New York, NY,
  USA (2022). \doi{10.1145/3477495.3531879},
  \url{https://doi.org/10.1145/3477495.3531879}

\bibitem{alpaca}
Taori, R., Gulrajani, I., Zhang, T., Dubois, Y., Li, X., Guestrin, C., Liang,
  P., Hashimoto, T.B.: Stanford alpaca: An instruction-following llama model
  (2023)

\bibitem{LLaMA}
Touvron, H., Lavril, T., Izacard, G., Martinet, X., Lachaux, M.A., Lacroix, T.,
  Rozière, B., Goyal, N., Hambro, E., Azhar, F., Rodriguez, A., Joulin, A.,
  Grave, E., Lample, G.: Llama: Open and efficient foundation language models
  (02 2023). \doi{10.48550/arXiv.2302.13971}

\bibitem{vaswani2017attention}
Vaswani, A., Shazeer, N., Parmar, N., Uszkoreit, J., Jones, L., Gomez, A.N.,
  Kaiser, {\L}., Polosukhin, I.: Attention is all you need. Advances in neural
  information processing systems  \textbf{30} (2017)

\bibitem{wei2023chainofthought}
Wei, J., Wang, X., Schuurmans, D., Bosma, M., Ichter, B., Xia, F., Chi, E., Le,
  Q., Zhou, D.: Chain-of-thought prompting elicits reasoning in large language
  models (2023)

\bibitem{wu2021towards}
Wu, J.Y., Shieh, A.T.W., Hsu, S.J., Chen, Y.N.: Towards generating citation
  sentences for multiple references with intent control. arXiv preprint
  arXiv:2112.01332  (2021)

\bibitem{xing2020automatic}
Xing, X., Fan, X., Wan, X.: Automatic generation of citation texts in scholarly
  papers: A pilot study. In: Proceedings of the 58th Annual Meeting of the
  Association for Computational Linguistics. pp. 6181--6190 (2020)

\bibitem{ye2022plmarker}
Ye, D., Lin, Y., Li, P., Sun, M.: Packed levitated marker for entity and
  relation extraction. In: Muresan, S., Nakov, P., Villavicencio, A. (eds.)
  Proceedings of the 60th Annual Meeting of the Association for Computational
  Linguistics (Volume 1: Long Papers), {ACL} 2022, Dublin, Ireland, May 22-27,
  2022. pp. 4904--4917. Association for Computational Linguistics (2022),
  \url{https://aclanthology.org/2022.acl-long.337}

\bibitem{ye2023incontext}
Ye, S., Hwang, H., Yang, S., Yun, H., Kim, Y., Seo, M.: In-context instruction
  learning (2023)

\bibitem{yeloglu2011multi}
Yeloglu, O., Milios, E., Zincir-Heywood, N.: Multi-document summarization of
  scientific corpora. In: Proceedings of the 2011 ACM Symposium on Applied
  Computing. pp. 252--258 (2011)

\end{thebibliography}

\appendix
\section{Appendix}

This section shows the inference examples used to test the fine-tuned model and checking the generated text quality, and context.

\begin{figure}[ht]
\centering
  \includegraphics[width = 1\linewidth]{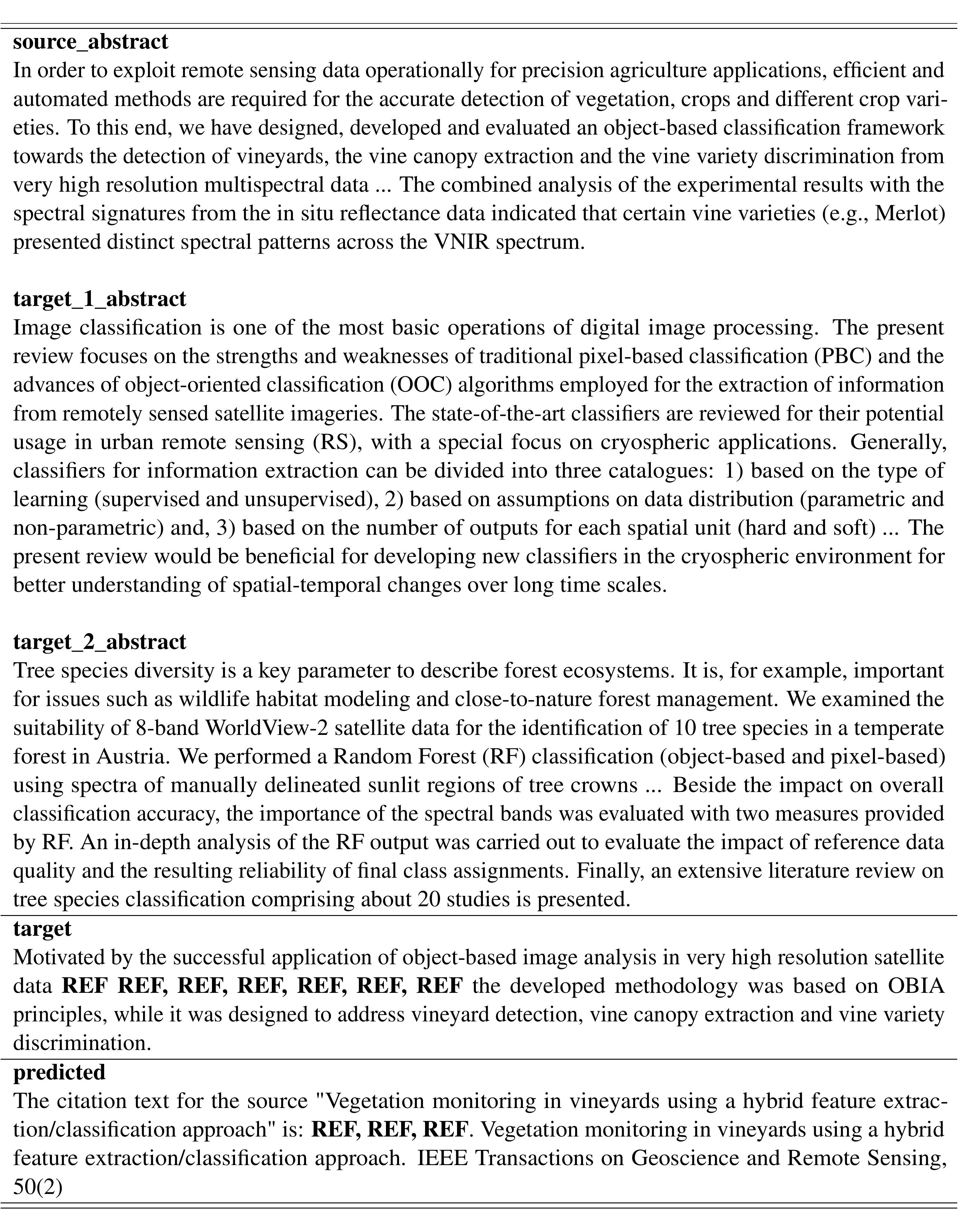}
\caption{Example of Generated Citation text from the best Model \textbf{(Vicuna)} without knowledge graph relations}
\label{example1}
\end{figure}

\begin{figure*}[ht]
\centering
  \includegraphics[width = 1\linewidth]{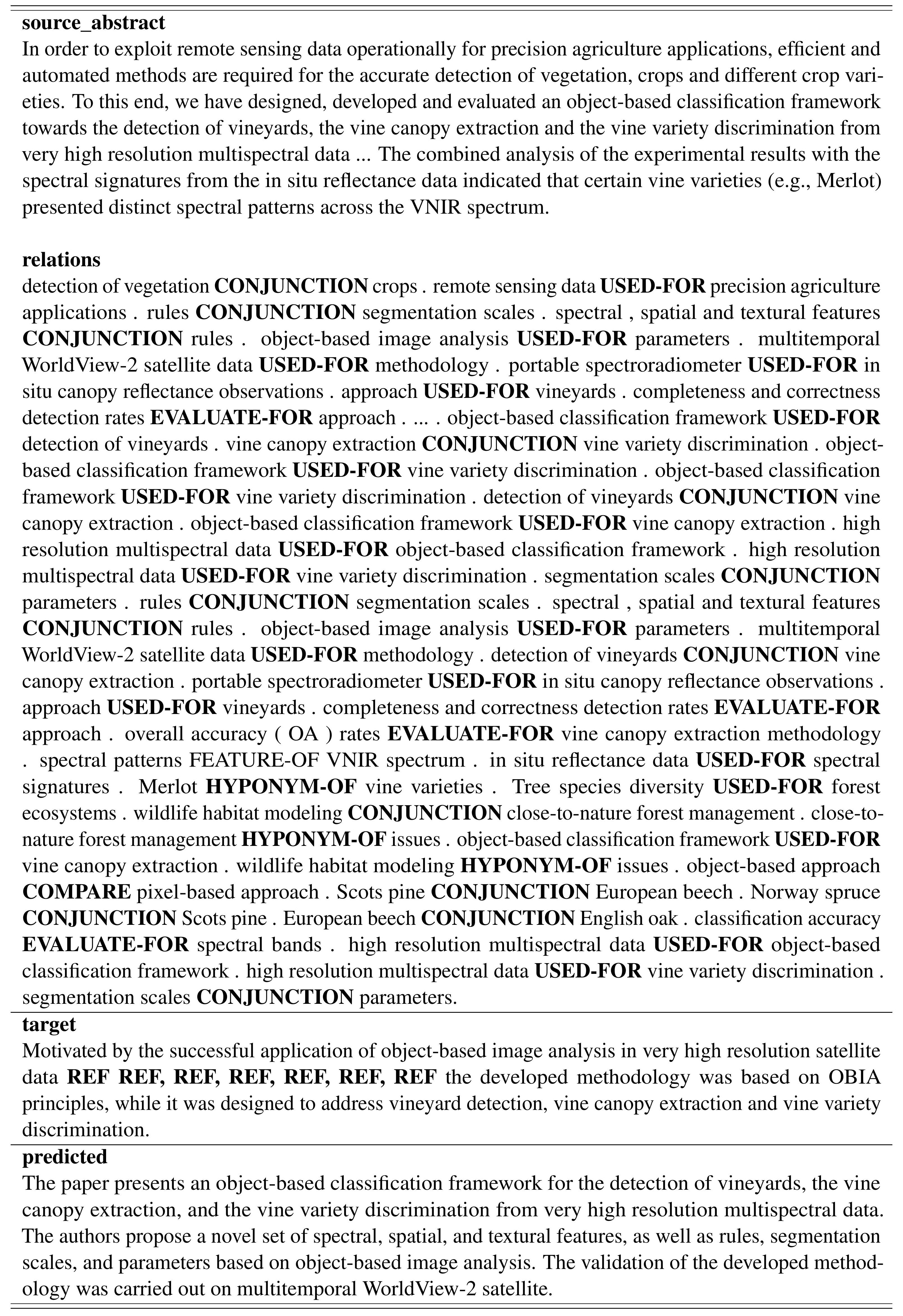}
\caption{Example of Generated Citation text from the best Model \textbf{(Vicuna)} with knowledge graph relations}
\label{example2}
\end{figure*}

A sample of the generated citation text using the improved Vicuna model is shown in Figure.~\ref{example1}. With the help of the supplied source\_abstract and set of target\_abstracts, high-quality generated citation text that fits both the source and target papers contexts was produced. Due to the inclusion of knowledge graph relations, the generated citation text in Figure.~\ref{example2} exhibits a higher level of context richness. These linkages help generate text that is more contextually relevant by improving our knowledge of the relationships between words in the source and target abstracts.

\end{document}